\documentclass[conference,a4paper]{IEEEtran}
\usepackage{amsmath,graphicx}
\usepackage{times}
\usepackage{epsfig}
\usepackage{amssymb}
\usepackage{nicefrac}
\usepackage{epstopdf}
\usepackage{booktabs}
\usepackage{multirow}
\usepackage[table]{xcolor}
\usepackage{caption}
\usepackage{subcaption}
\usepackage{dblfloatfix}
\usepackage[linesnumbered,ruled]{algorithm2e}
\usepackage{algpseudocode}

\newcommand\blfootnote[1]{%
  \begingroup
  \renewcommand\thefootnote{}\footnote{#1}%
  \addtocounter{footnote}{-1}%
  \endgroup
}
\usepackage{multirow}
\usepackage{bm}
\usepackage{slashbox}
\usepackage{bm}
\usepackage[
linkcolor=red,
anchorcolor=blue,
citecolor=green]{hyperref}
\begin{document}
\title{Deep joint rain and haze removal\\from single images}
\author{\IEEEauthorblockN{Liang Shen$^*$, Zihan Yue$^*$, Quan Chen, Fan Feng and Jie Ma}
\IEEEauthorblockA{Institute of Image Recognition and Artificial Intelligence\\
School of Automation, Huazhong University of Science and Technology, China
\\\texttt{$\{$shenliang,yuezihan,chenquan,fengfan,majie$\}$@hust.edu.cn}}}
\maketitle

\begin{abstract}
Rain removal from a single image is a challenge which has been studied for a long time. In this paper, a novel convolutional neural network based on wavelet and dark channel is proposed. On one hand, we think that rain streaks correspond to high frequency component of the image. Therefore, haar wavelet transform is a good choice to separate the rain streaks and background to some extent. More specifically, the LL subband of a rain image is more inclined to express the background information, while LH, HL, HH subband tend to represent the rain streaks and the edges. On the other hand, the accumulation of rain streaks from long distance makes the rain image look like haze veil. We extract dark channel of rain image as a feature map in network. By increasing this mapping between the dark channel of input and output images, we achieve haze removal in an indirect way. All of the parameters are optimized by back-propagation. Experiments on both synthetic and real-world datasets reveal that our method outperforms other state-of-the-art methods from a qualitative and quantitative perspective.
\end{abstract}

\IEEEpeerreviewmaketitle
\section{Introduction}

Most computer vision tasks assume the sufficient high-quality of images. However, various degradations often occur in realistic scenes. For example, rainy weather becomes an inevitable situation when these tasks are applied to outdoor scenes. The rain in image can be roughly divided into two cases. Rain streaks near to the camera lens can be considered as noise in the image, whereas rain from long distance looks like fog.\blfootnote{$^*$ indicates equal contribution by authors.}

In this paper, a novel convolutional neural network based on wavelet~\cite{Vidakovic1999An} and dark channel~\cite{he2011single} has been proposed to remove different types of rain. On one hand, considering that rain streaks mainly correspond to the high frequency components in the image, we think that the wavelet-based approach is probably a good choice. Firstly, the rain image and the ground truth are transformed into four sub-images (low-low, low-high, high-low, high-high frequency) by using Haar wavelet~\cite{Vidakovic1999An} respectively. Then we try to train an end-to-end mapping between these different sub-images in wavelet domain to remove the light rain. On the other hand, the accumulation of rain streaks from long distance makes the image overcast as if covered by haze. In this condition, the dark channel prior proposed by He \textit{et al.}\cite{he2011single} can still be considered as a good approach to remove the veil from an image. However, in this model we regard dark channel as a feature map in convolutional neural network. By combining above two different methods in a consistent framework, the final model is considered as a multi-task optimization problem and all parameters are optimized by back-propagation.

\section{Related works}

Similar to many image restoration problems such as image denoising~\cite{liu2017image}, image deblurring~\cite{vasiljevic2016examining}, super resolution~\cite{dong2016image} and image dehazing~\cite{cai2016dehazenet}, image deraining is also a noticeable field. The common point of these methods is to solve the inverse problem by using the degraded images. As we all know, inverse problems are the pathological problems with infinite solutions. To overcome this difficulty, many prior-based and regularization methods have been proposed. In particular, rain removal can be divided into two groups: Video based methods and single image based methods.

\subsection{Video based methods}

Removing rain from video has been widely explored. Kshitiz \textit{et al.}\cite{garg2005does} analysed the visual effects of rain on an imaging system. They developed a physics-based blur model that explained the photometry of rain. Barnum \textit{et al.}\cite{barnum2010analysis} studied the phenomenon of rain in frequency domain. They revealed that dynamic weathers such as rain and snow have a significant effect in frequency space. Bossu \textit{et al.}\cite{bossu2011rain} separated the foreground from background in image sequences by using a classical Gaussian mixture model. The histogram of orientation of rain streaks maked it possible to detect the pixel of rain in the foreground image. Chen \textit{et al.}\cite{chen2013generalized} proposed a novel low-rank model from matrix to tensor structure to capture the correlated rain streaks. Recently, Jiang \textit{et al.}\cite{jiang2017novel} proposed a novel tensor-based approach by considering the inherent property of rain streaks and cleared the videos. All of these methods make full use of temporal in adjacent frames to figure out rain streaks in video.

\subsection{Single image methods}

Compared with multi-frame rain removal, single image deraining is more difficult due to the lack of temporal information. Traditional methods are usually based on image decomposition, sparse coding or dictionary learning. For example, Fu \textit{et al.}\cite{fu2011single} treated the rain removal as an image decomposition problem by using morphological component analysis. Li \textit{et al.}\cite{li2016rain} tried to use simple patch-based priors for both foreground and background. Xu \textit{et al.}\cite{xu2012improved} used filtering method to remove rain by guiding image such as guider filter proposed by He \textit{et al.}\cite{he2011single}. In~\cite{luo2015removing}, Luo proposed a dictionary learning method for single image deraining. Besides these, deep learning makes a great achievement in many low-level vision tasks. Dong \textit{et al.}\cite{dong2016image} attempted to use convolutional neural network in image super-resolution for the first time and achieved remarkable improvement. After this, a large number of similar methods spewed out. For instance, Chakrabarti and Ayan~\cite{chakrabarti2016neural} proposed a novel neural approach for blind motion deblurring which uses the trained network to compute sharp image patches. Cai \textit{et al.}\cite{cai2016dehazenet} proposed an end-to-end system named dehazeNet which is used to estimate the medium transmission. Fu \textit{et al.}\cite{Fu2017Removing} directly learned the mapping between rain image and high-frequency detail image by using the residual structure proposed in~\cite{he2016deep}. Due to the fact that rain streaks removal in an image is almost an identity mapping, residual structure will make the learning process easier. Yang \textit{et al.}\cite{yang2016joint} constructed a multi-task that solved the inverse problem through an end-to-end learning. They also proposed a novel network for extracting the rain discriminative feature to leverage more content.

\section{Proposed method}

\begin{figure}[t]
\centering
        \begin{subfigure}[b]{0.22\textwidth}
            \includegraphics[width=\textwidth]{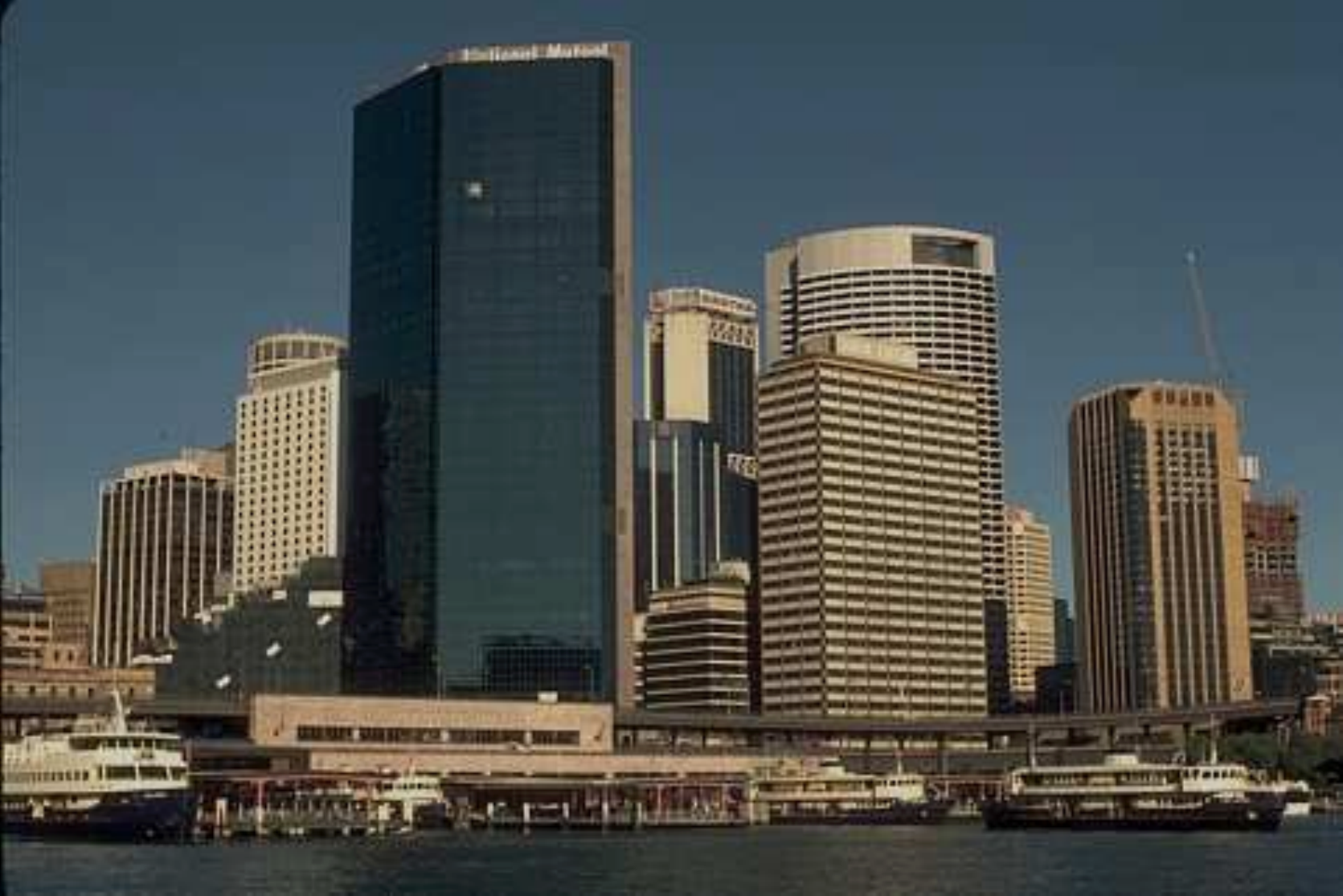}
            \caption{Ground truth}
        \end{subfigure}
        \begin{subfigure}[b]{0.22\textwidth}
            \includegraphics[width=\textwidth]{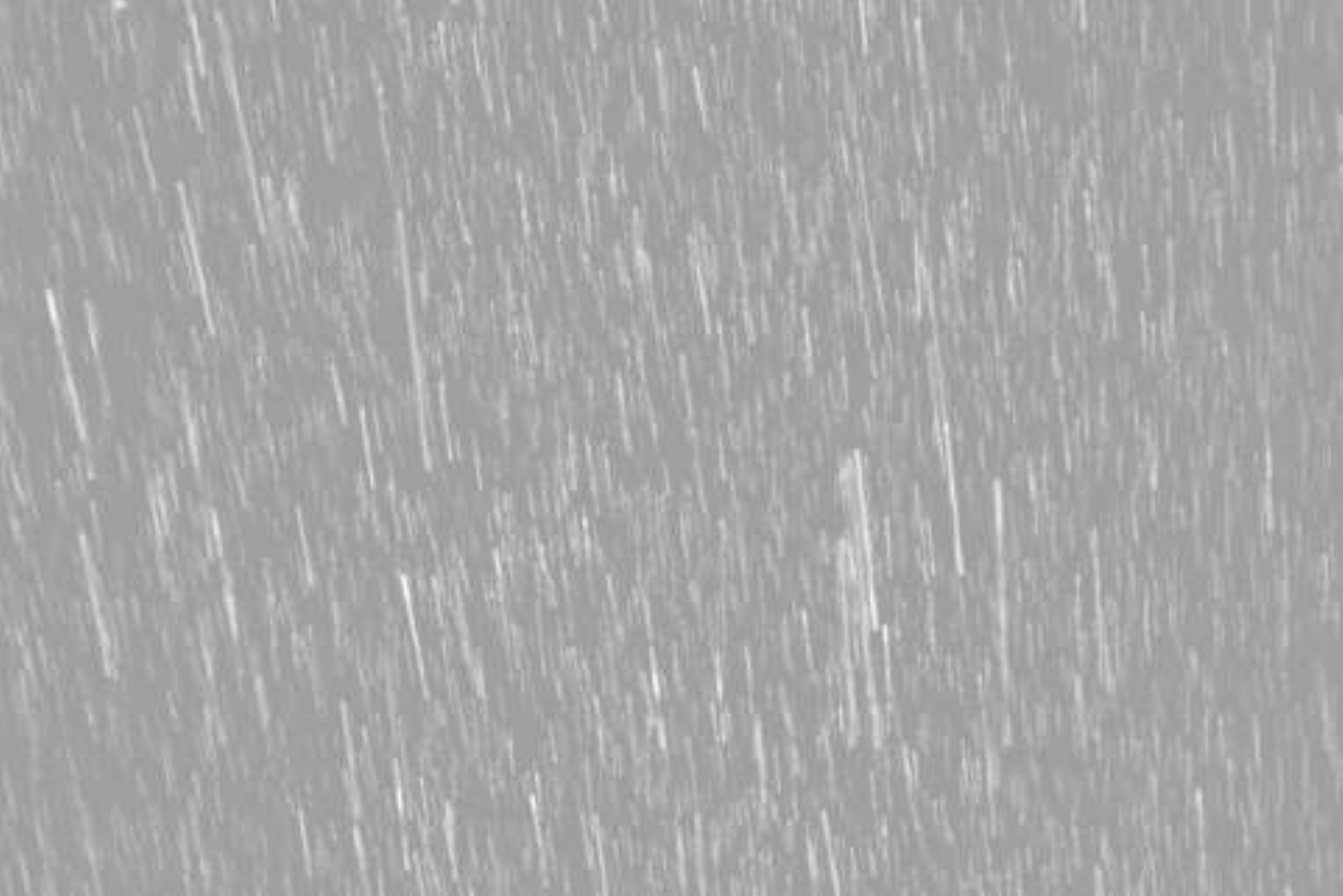}
            \caption{Rain streaks}
        \end{subfigure}
        \begin{subfigure}[b]{0.22\textwidth}
            \includegraphics[width=\textwidth]{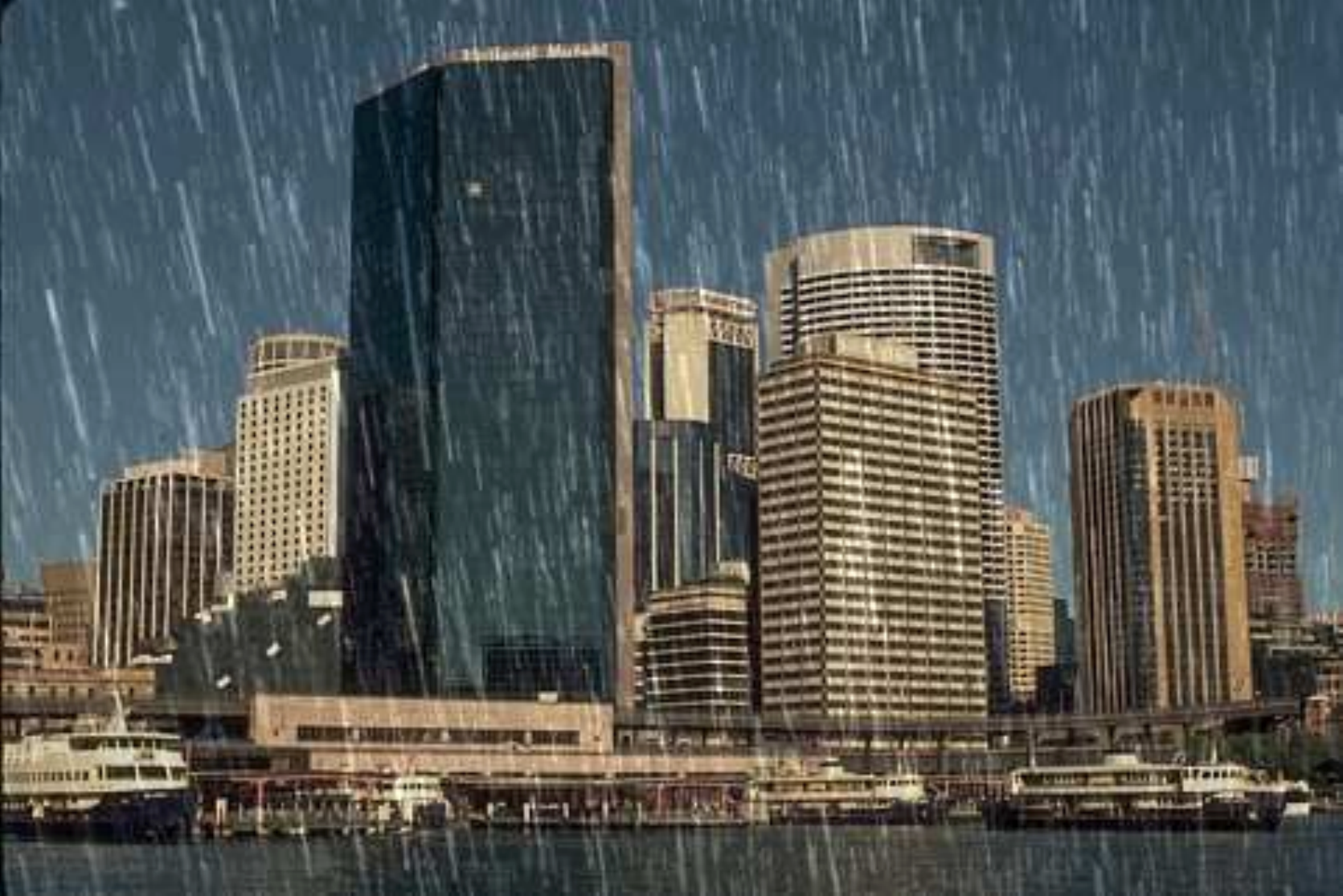}
            \caption{Rain image}
        \end{subfigure}
        \begin{subfigure}[b]{0.22\textwidth}
            \includegraphics[width=\textwidth]{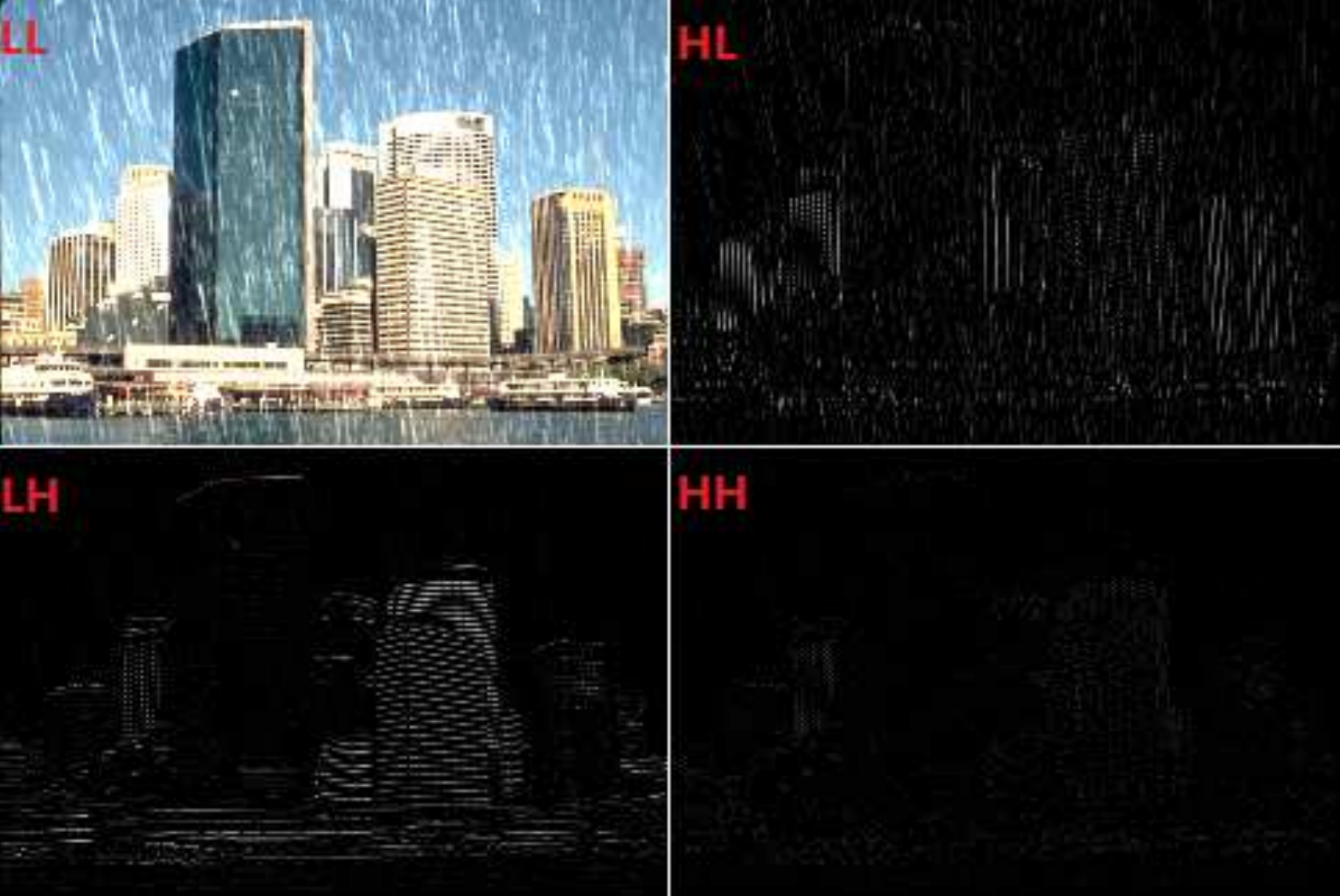}
            \caption{The wavelet result}
        \end{subfigure}
        \caption{(a)ground truth,(b)image with rain streaks,(c) rain image formed by (a) and (b).(d) are the four sub-images of haar wavelet transform result. Obviously, HL subband contains more raindrops due to the fact that rain is falling down from the top, by the contrast, LH subband contains more the ground truth edge.(Best zoom in the LH and HL subband on screen)}
        \label{fig:haarwavelet}
\end{figure}

In this section, we elaborate that rain in image can be roughly divided into two situations. Rain streaks near to the lens look like noise in an image, whereas rain from distance looks like haze veil. Our model takes the above two aspects into account. At last, by combining above two separate structures to one network, the final model can be considered as an end-to-end structure for rain and haze removal.

\subsection{Rain model in an image}

Traditional rain model is composed of two components: rain streaks and background. Mathematically, it can be expressed as:
\begin{equation}\label{equ:baseequation}
O = B + R
\end{equation}

Where $O$ represents the observed degraded image, $B$ is the background scene and $R$ are the rain streaks. However, in many cases where the accumulation of rain streaks from long distance makes the image overcast as if covered by haze veil, which causes the model perform not good enough. Based on this baseline, many modified models have been proposed. For example, considering the dense rain and fog phenomenon, Yang \textit{et al.}\cite{yang2016joint} extended Equation~\ref{equ:baseequation} to create a new model to accommodate them:
\begin{equation}
O = \alpha (B + \sum\limits_{t = 1}^s {{R_t}} ) + (1 - \alpha )A
\end{equation}

As we can see, the first item of above equation consists of several different layers of rain streaks. $A$ and $\alpha$ in the second item are global atmospheric light and scene transmission respectively, which have been described in haze removal papers such as~\cite{he2011single}.

\subsection{Wavelet for rain streaks removal}

 Fourier transform is a nice tool for analysing images in the frequency domain, however, the spectrum of an image loses a lot of great properties such as local receptive field, which makes it difficult to use convolutional neural network. Fortunately, the other frequently-used method, called wavelet transform, is now making it easier to analyse the images. Different from Fourier transform based on sinusoids, wavelet transform is based on small waves, which is more convenient to train. In this paper, we attempt to use one of the most commonly used wavelet: Haar wavelet.

\begin{figure}[t]
  \centering
  \includegraphics[width=0.48\textwidth]{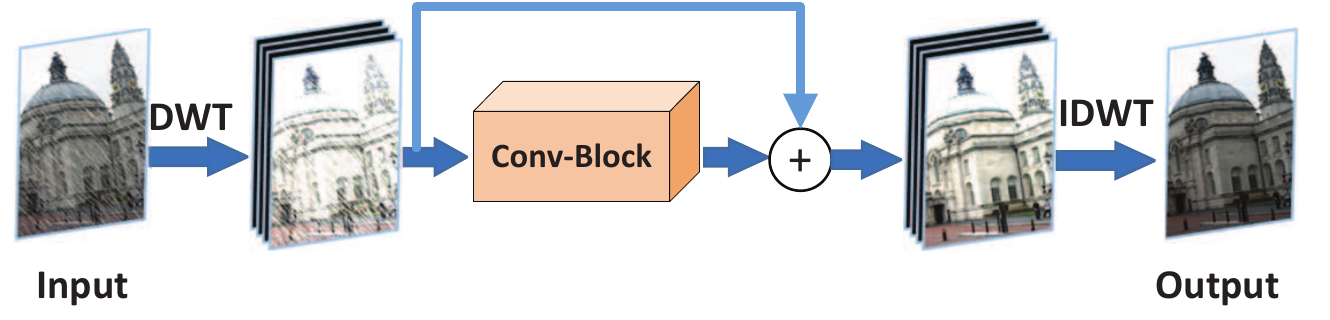}\\
  \caption{The simple rain removal network architecture(SRR-net) for removing the rain image without the haze veil}\label{model1}
\end{figure}

\begin{figure*}[t]
  \centering
  \includegraphics[width=0.8\textwidth]{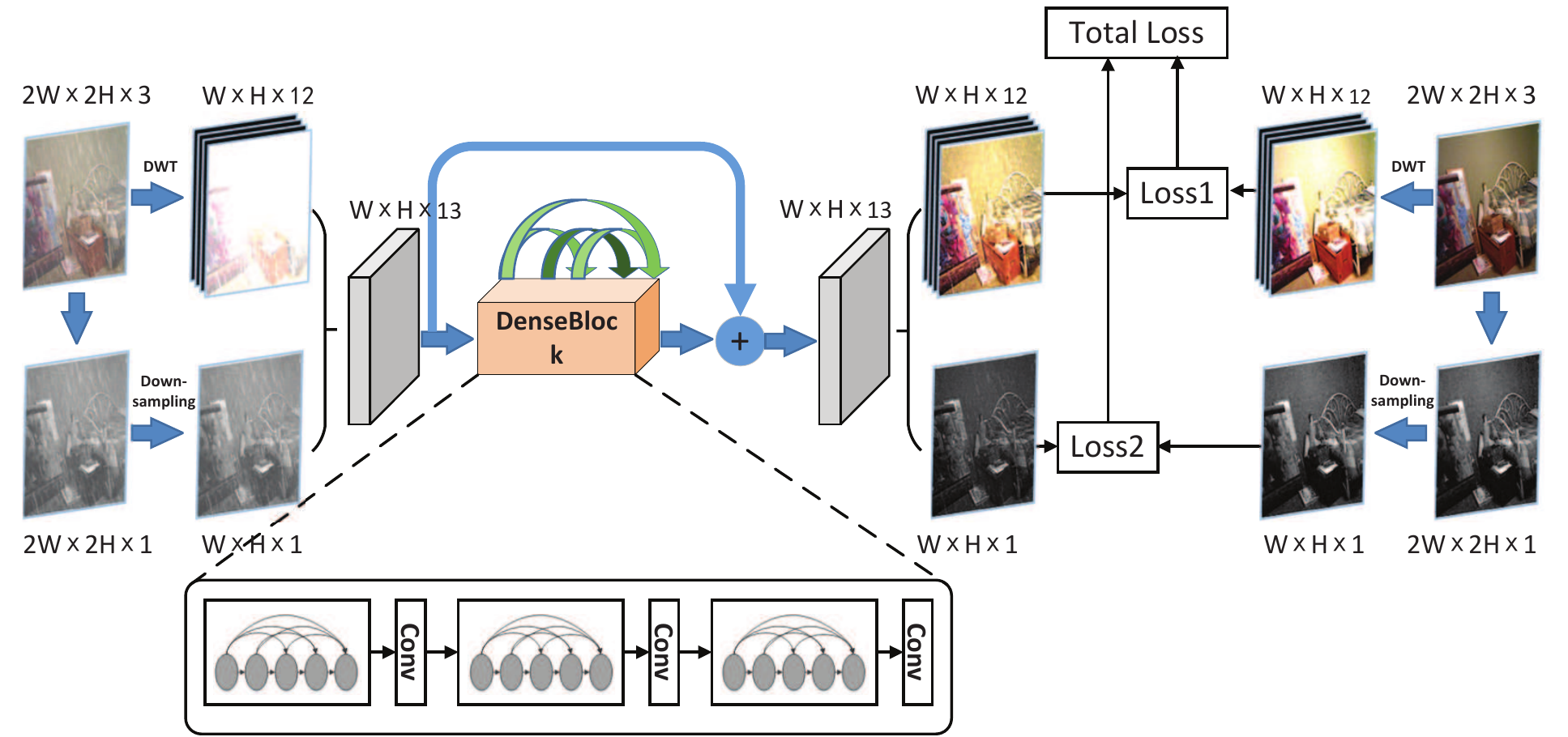}\\
  \caption{Deep joint rain and haze removal network(DJRHR-net), which is training for removing the rain and the haze veil.}\label{model2}
\end{figure*}

Figure~\ref{fig:haarwavelet} gives an example of discrete wavelet transform using Haar basis function. As we can see, the four sub-images correspond to the approximation subband LL, horizontal detail LH, vertical detail HL and diagonal detail HH, respectively. LL subband represents the main content of an image, whereas LH, HL and HH represent the detail information of an image. More specifically, the LL subband of a rain image is more inclined to express the background information, and HL subband contains more raindrops due to the fact that the rain is falling down from the top, while the LH subband includes the more edge information of the background. Therefore, this decomposition is not only helpful to get rid of the rain noise but also to protect the edge details.

The goal of rain removal is to recover low-quality(LQ) images to high-quality(HQ) images. Considering that wavelet subbands can better represent the shape of rain streaks and edges, our network tries to fit these subbands. Firstly, we convert original images (LQ, HQ) to wavelet domain, which are considered as input image and train label respectively:

\begin{equation}
\begin{array}{*{20}{c}}
X:\left\{ {L{Q_{LL}},L{Q_{LH}},L{Q_{HL}},L{Q_{HH}}} \right\} = DWT(LQ)\\
Y:\left\{ {H{Q_{LL}},H{Q_{LH}},H{Q_{HL}},H{Q_{HH}}} \right\} = DWT(HQ)
\end{array}
\end{equation}

Inspired by the residual structure~\cite{he2016deep}, Fu \textit{et al.}\cite{Fu2017Removing} proposed an end-to-end network between rain image and high-frequency detail image. Compared with their method which directly learns the mapping between original images, our model attempts to fit the mapping between these sub-images generated by wavelet as shown in Figure~\ref{model1}. Mathematically, the loss function of SRR-net can be defined as:

\begin{equation}
L = \frac{1}{N}\sum\limits_{i = 1}^N {\left\| {{Y_i} - {X_i} - f({X_i})} \right\|_F^2}  + \lambda \left\| W \right\|_F^2
\end{equation}

Where $X_i$ and $Y_i$ represent the tensors that concatenate four wavelet subbands of LQ and HQ images respectively, $N$ is equal to the number of the images on training dataset. $f( \cdot )$ represents the nonlinear mapping of the neural network, and in this paper we use ordinary convolution and ReLU layers. $W$ corresponds to the parameters of the whole model which are optimized by back-propagation, and ${\left\|  \cdot  \right\|_F}$ represents the Frobenius norm. More experimental results and training parameter setting will be elaborated in section~\ref{experiments}.

To summmarize, after the whole network(SRR-net) is trained, the whole rain removal process is as follows:

Firstly, we convert the rain image to wavelet domain by Haar wavelet and concatenate these four subbands to a tensor $X$ with 12 channels.
\begin{equation}
X:\left\{ {L{Q_{LL}},L{Q_{LH}},L{Q_{HL}},L{Q_{HH}}} \right\} = DWT(LQ)
\end{equation}

Next, we put the wavelet subbands $X$ to the trained residual network.

\begin{equation}
Y = f(X) + X
\end{equation}

At last, the inverse wavelet transform is used to generate final high-quality result.
\begin{equation}
HQ = IDWT(Y)
\end{equation}

\subsection{Dark channels for rain accumulation}

\begin{figure*}[t]
\centering
        \begin{subfigure}[b]{0.13\textwidth}
            \includegraphics[width=\textwidth]{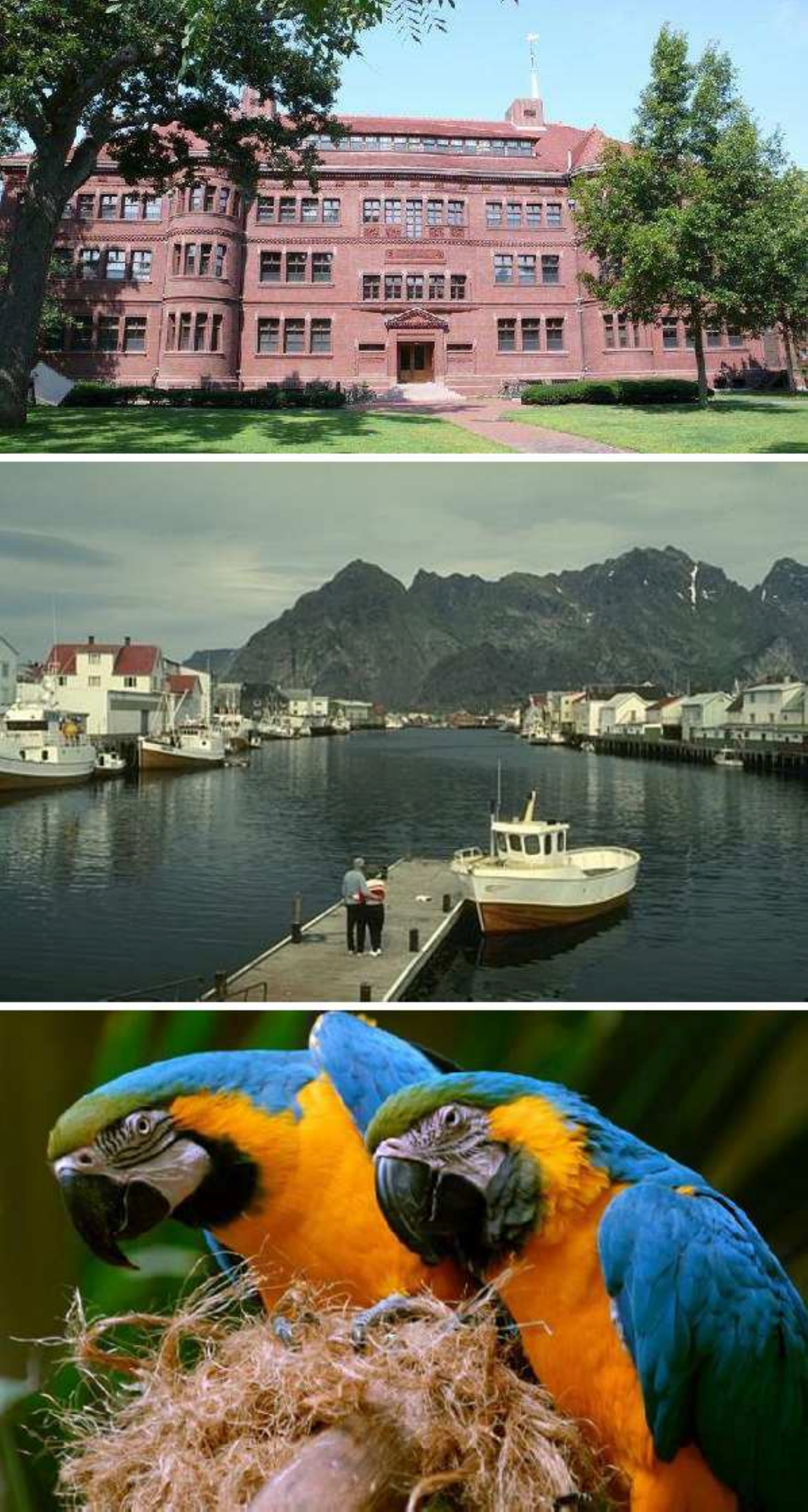}
            \caption{Ground truth}
        \end{subfigure}
        \begin{subfigure}[b]{0.13\textwidth}
            \includegraphics[width=\textwidth]{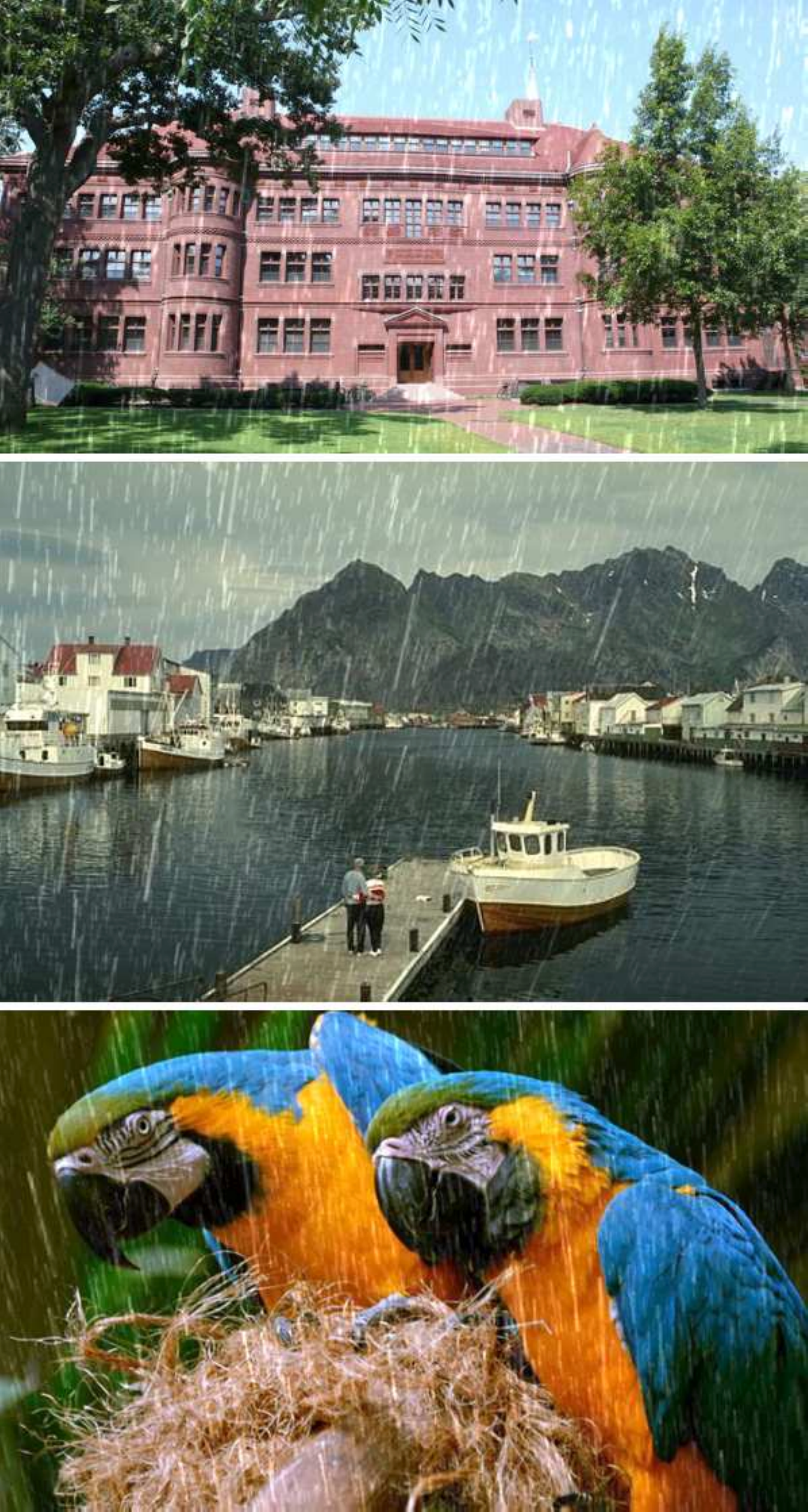}
            \caption{Synthetic data}
        \end{subfigure}
        \begin{subfigure}[b]{0.13\textwidth}
            \includegraphics[width=\textwidth]{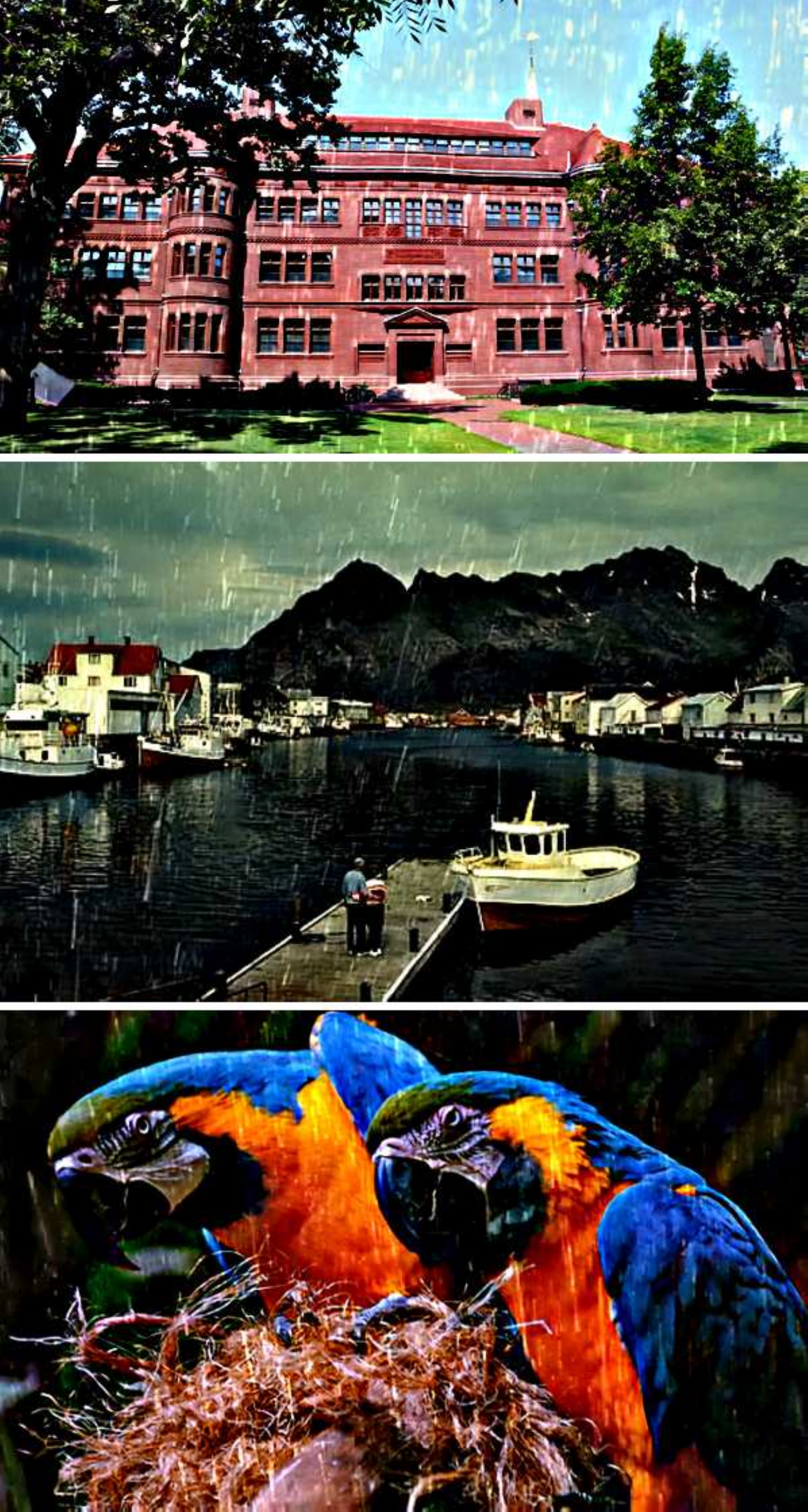}
            \caption{SIRR-net\cite{Fu2016Clearing}}
        \end{subfigure}
        \begin{subfigure}[b]{0.13\textwidth}
            \includegraphics[width=\textwidth]{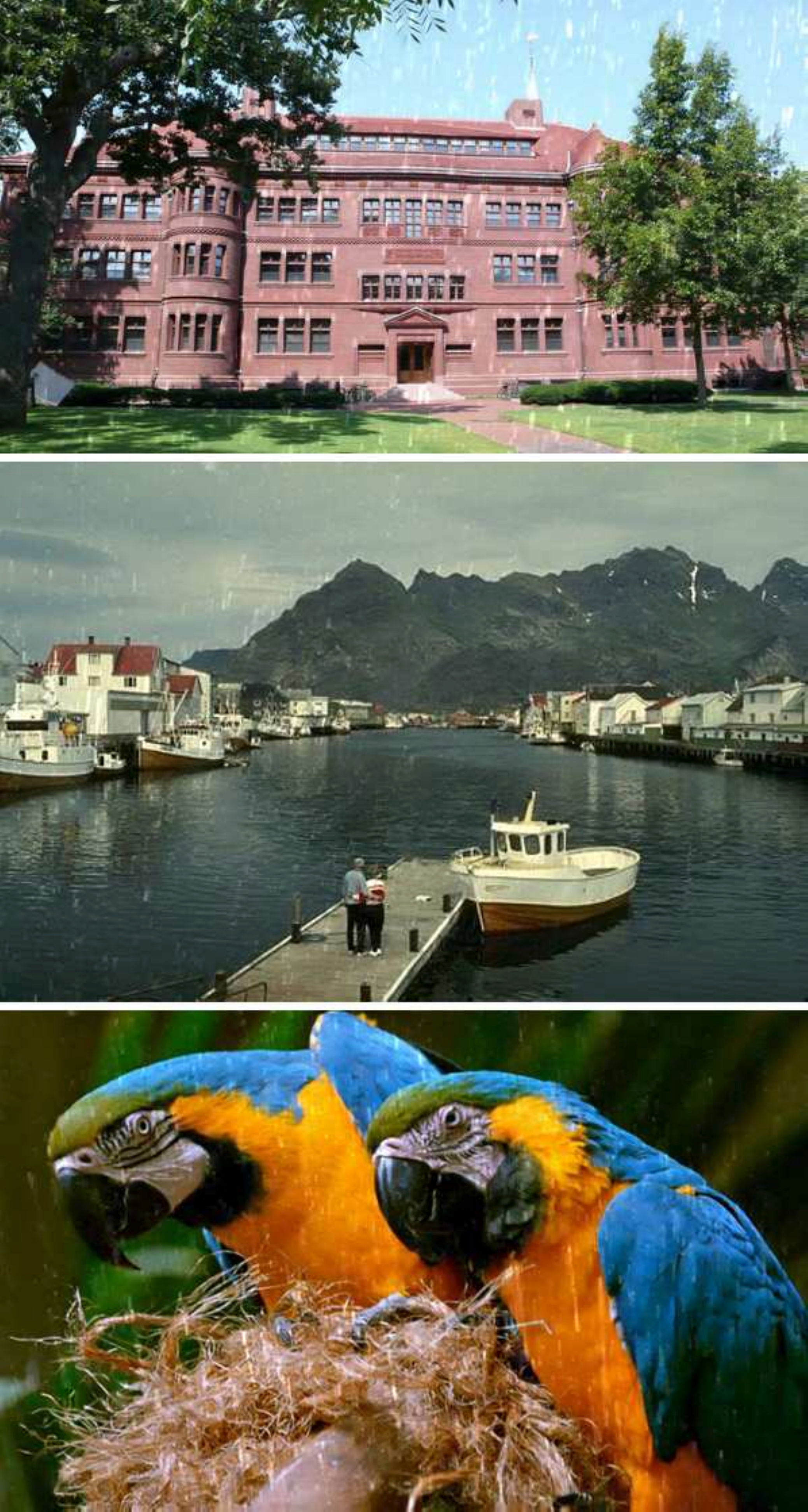}
            \caption{Detail-net\cite{Fu2017Removing}}
        \end{subfigure}
        \begin{subfigure}[b]{0.13\textwidth}
            \includegraphics[width=\textwidth]{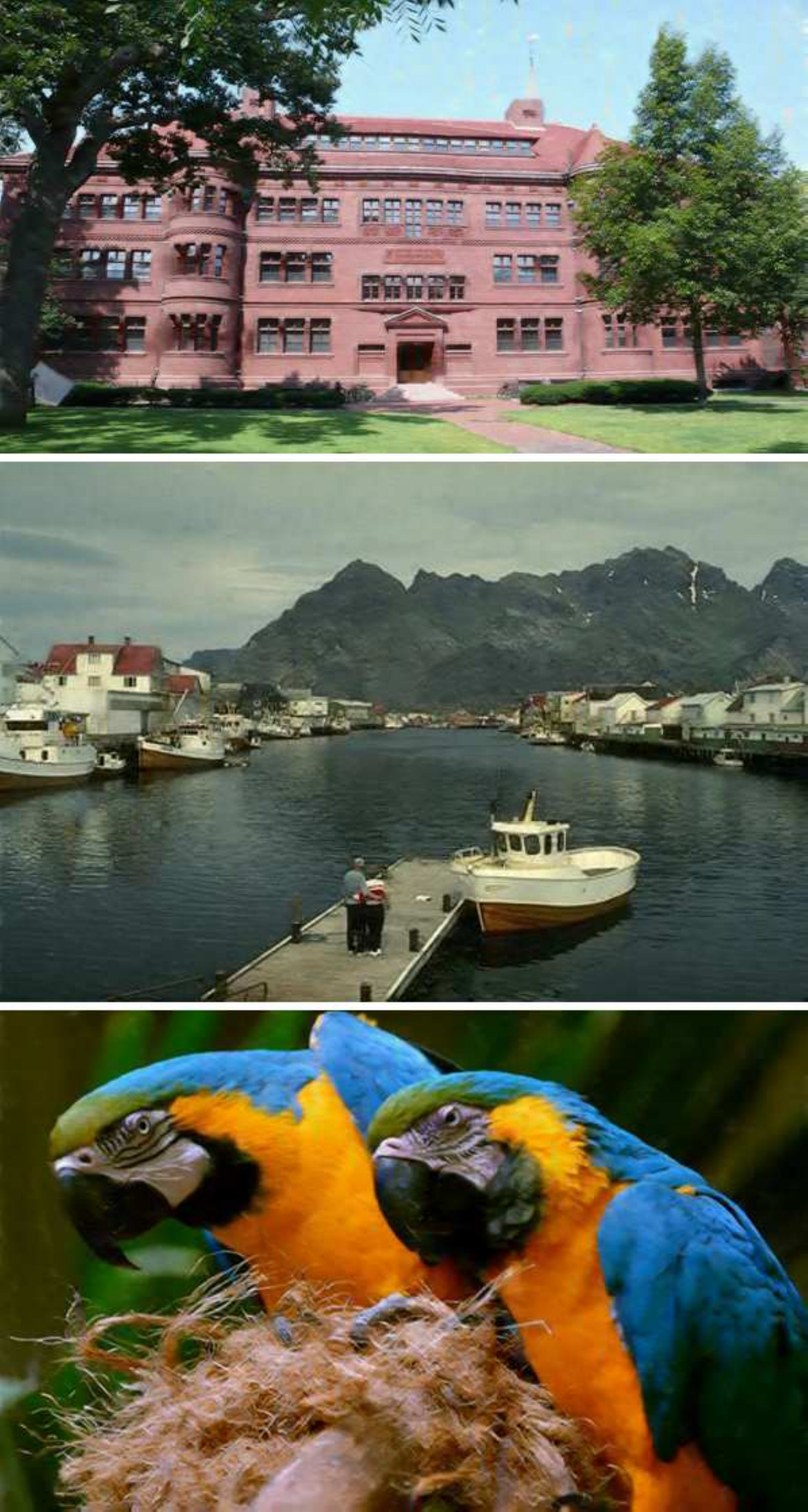}
            \caption{JORDER\cite{yang2016joint}}
        \end{subfigure}
        \begin{subfigure}[b]{0.13\textwidth}
            \includegraphics[width=\textwidth]{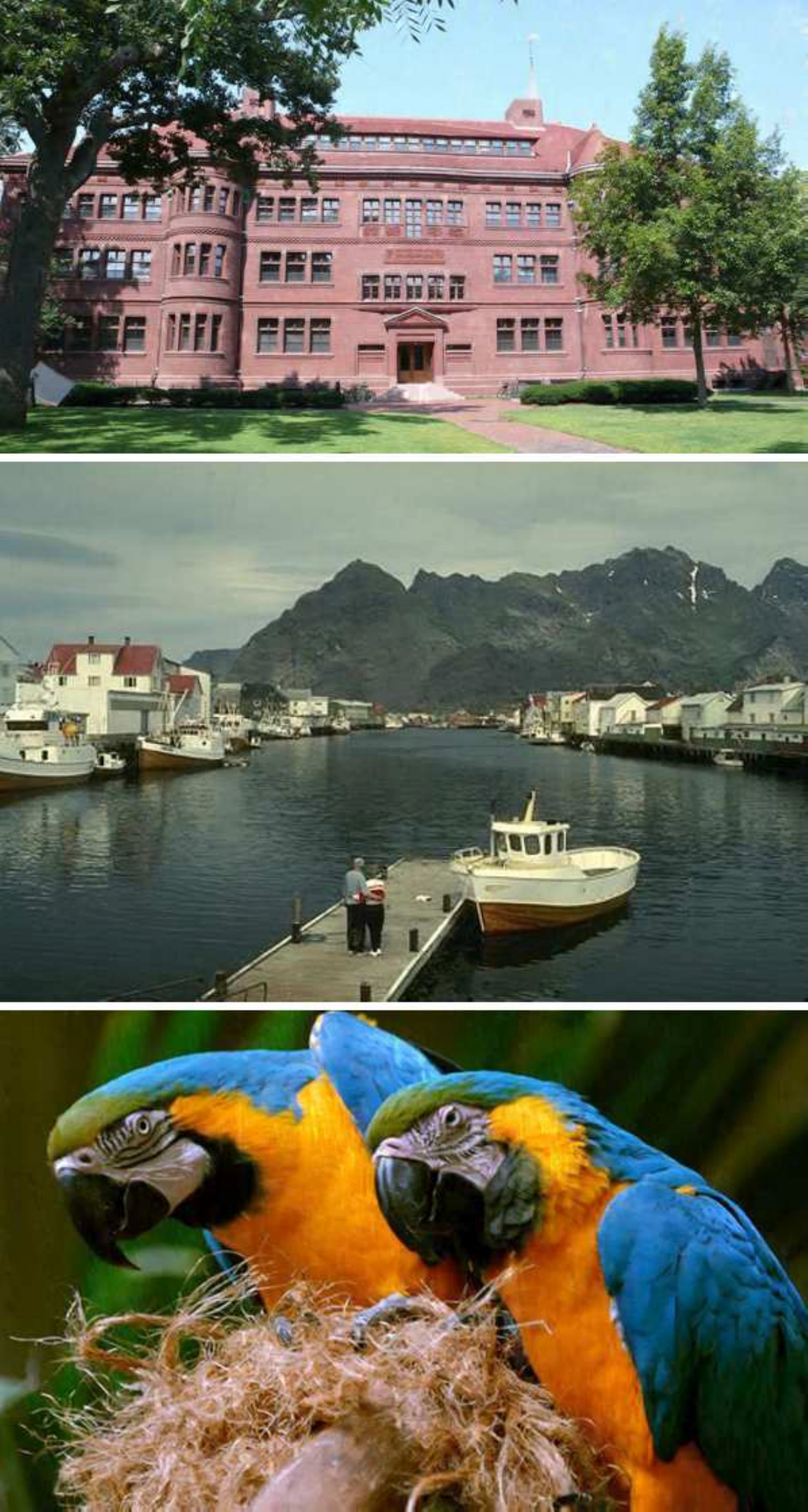}
            \caption{Our SRR-net}
        \end{subfigure}
        \caption{Results using different methods on synthesized test images}
        \label{fig:synthesized}
\end{figure*}

\setlength{\tabcolsep}{5pt}
\begin{table*}[tbp]
\begin{center}
\caption{ Quantitative measurement results using PSNR/SSIM/NIQE on synthesized test images }
\begin{tabular}{| c || c c c| c c c |c c c| c c c| }
\hline
Method    & \multicolumn{3}{c|}{1st row}          & \multicolumn{3}{c|}{2nd row}           & \multicolumn{3}{c|}{3rd row}       &\multicolumn{3}{c|}{200 test images}     \\
\hline
Metric                         & PSNR   & SSIM  & NIQE    & PSNR   & SSIM  & NIQE  & PSNR   & SSIM  & NIQE   & PSNR   & SSIM  & NIQE\\
\hline
Synthetic image                &21.22  &0.72    &3.33       &27.13   &0.78   &4.07     &23.16   &0.65   &3.28     &28.91   &0.85   &4.14\\
\hline
 SIRR-net\cite{Fu2016Clearing} & 14.55  & 0.50 & 4.45       & 13.79 &0.44  & 5.47 & 15.84  & 0.43  & \textbf{2.80} & 13.76 &0.52 & 4.67\\
\hline
Detail-net\cite{Fu2017Removing}& 21.88  & 0.76 & \textbf{2.93}        & 31.00 & 0.92  & 3.53 & 25.03 & 0.77  & 2.87 & 29.13 & 0.92 & 3.29 \\
\hline
JORDER\cite{yang2016joint}      & 21.03  & 0.74 & 3.23        & 26.78 & 0.92  & 3.24 & 24.16 & 0.77  & 3.89 & 28.12 & 0.91 & 3.50 \\
\hline
\hline
Ours     & \textbf{22.53}   & \textbf{0.78} & 3.02 & \textbf{33.28} & \textbf{0.95}  & \textbf{2.78} & \textbf{26.87} & \textbf{0.83}  &2.85    & \textbf{30.19}  & \textbf{0.95}       & \textbf{3.18} \\
\hline
\hline
\end{tabular}
\label{table:SyntheticQuantitativeResult}
\end{center}
\end{table*}

In the previous section, we have explained the role of wavelet transform in deraining. However, the simple rain removal network can't handle the situation very well, where rain streaks are dense and these make the image have the haze veil. Haze removal is a traditional research direction which has been studied for a long time. One of the most classical methods is based on dark channel prior~\cite{he2011single}, which is a statistic of the haze-free images. By using this extra strong prior, the thickness of haze can be estimated and the high-quality image can be recovered directly.

In order to integrate the de-hazing into deep learning framework, we extract dark channel of an image as a feature map in convolutional neural network to contribute to the removal of this noise. It is more effective to add the artificial feature directly than the features learned by the deep network. So we increase a mapping between the dark channel of input and output images, which helps achieve haze removal through indirect means. Figure~\ref{model2} shows our final deep joint rain and haze removal network(DJRHR-net) which is designed as a multi-task architecture.

As we can see, the original LQ(rain image) and HQ(ground truth) should be converted into wavelet subbands and dark channel firstly.

\begin{figure*}[t]
\centering
        \begin{subfigure}[b]{0.18\textwidth}
            \includegraphics[width=\textwidth]{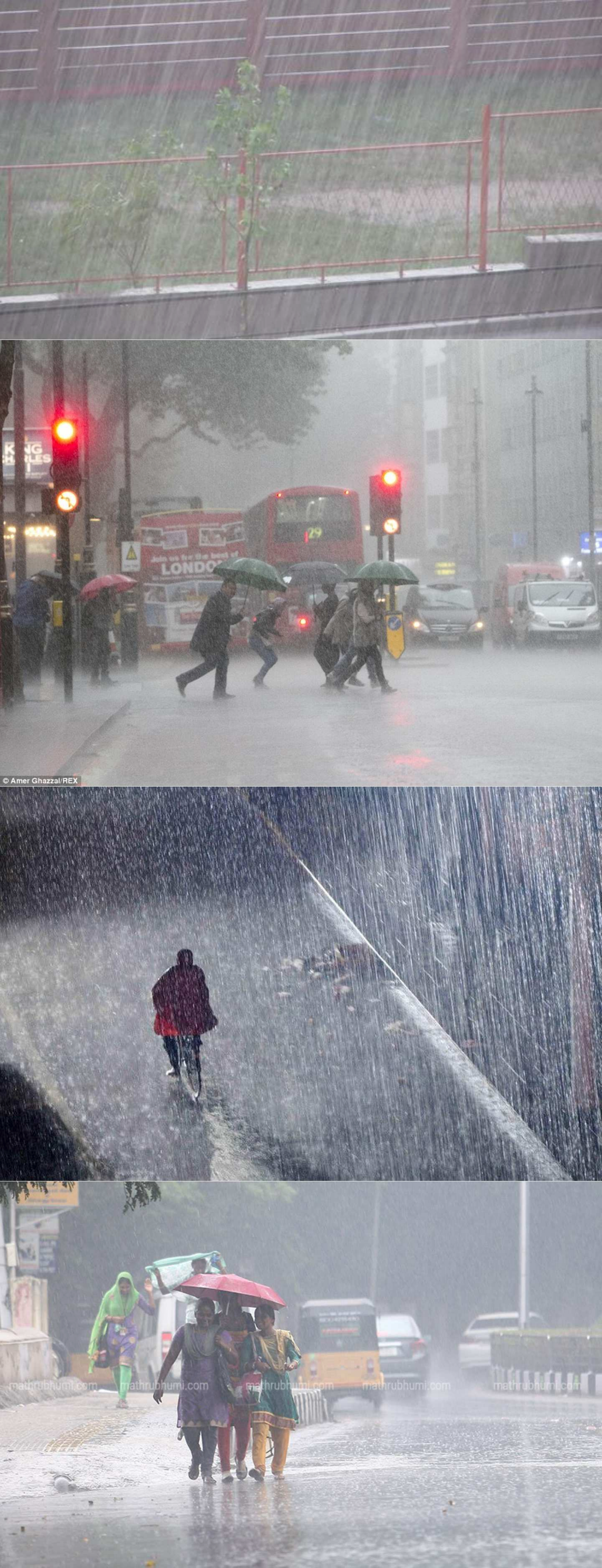}
            \caption{Real-world image}
        \end{subfigure}
        \begin{subfigure}[b]{0.18\textwidth}
            \includegraphics[width=\textwidth]{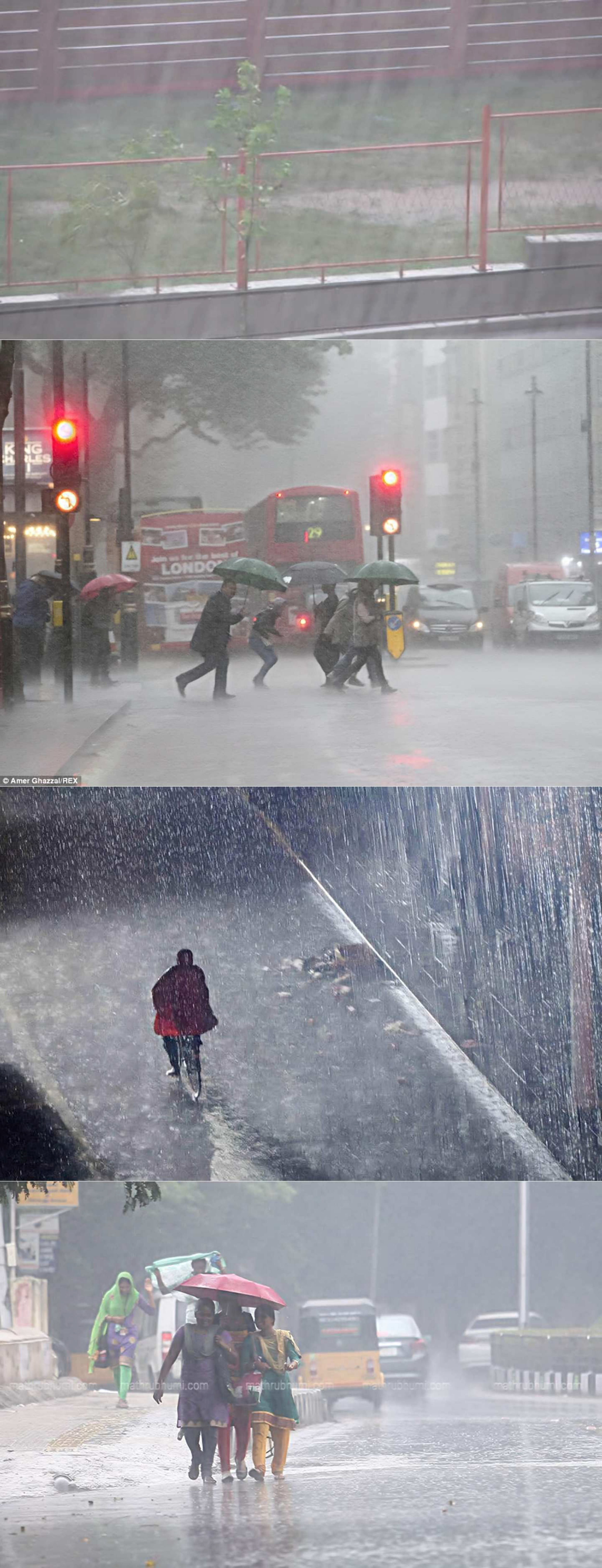}
            \caption{SIRR-net\cite{Fu2016Clearing}}
        \end{subfigure}
        \begin{subfigure}[b]{0.18\textwidth}
            \includegraphics[width=\textwidth]{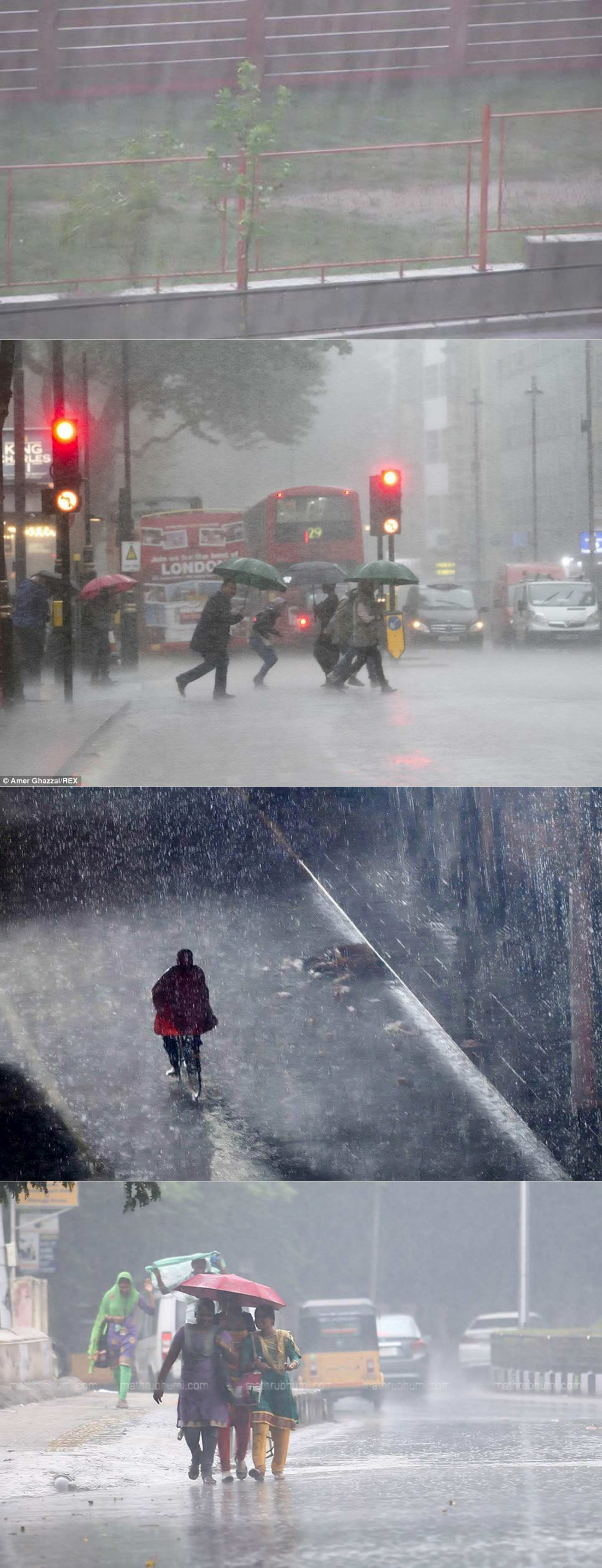}
            \caption{Detail-net\cite{Fu2017Removing}}
        \end{subfigure}
        \begin{subfigure}[b]{0.18\textwidth}
            \includegraphics[width=\textwidth]{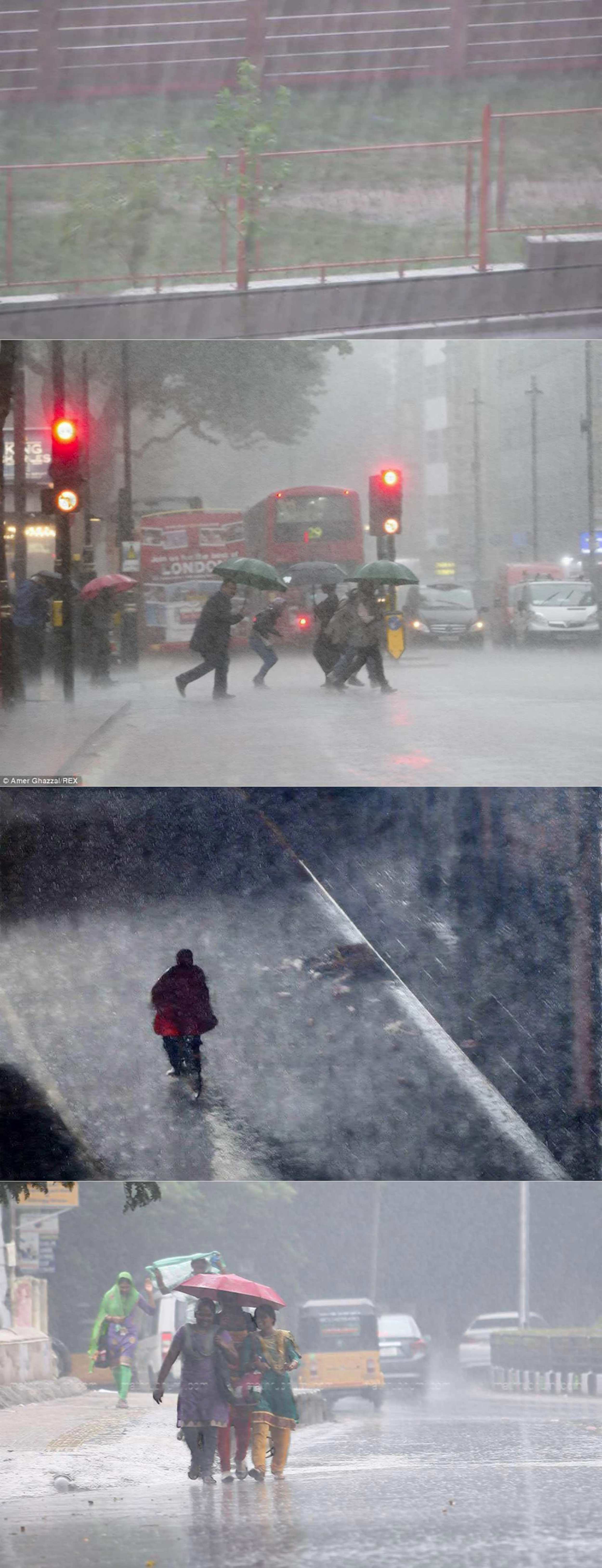}
            \caption{JORDER\cite{yang2016joint}}
        \end{subfigure}
        \begin{subfigure}[b]{0.18\textwidth}
            \includegraphics[width=\textwidth]{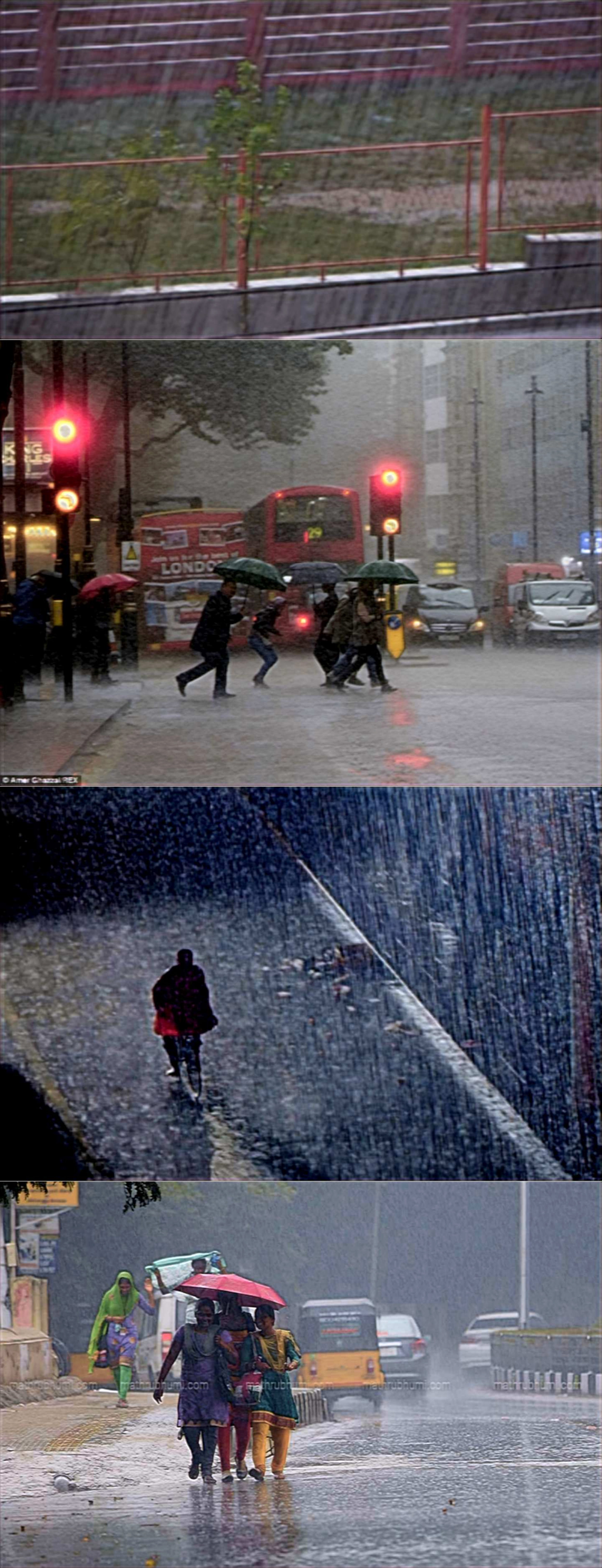}
            \caption{Our DJRHR-net}
        \end{subfigure}
        \caption{Results using different methods on real-world images}
        \label{fig:realworld}
\end{figure*}

\begin{equation}
\begin{array}{*{20}{c}}
{X:\left\{ {DWT(LQ),dark\_channel(LQ)} \right\}}\\
{Y:\left\{ {DWT(HQ),dark\_channel(HQ)} \right\}}
\end{array}
\end{equation}

Where
\begin{equation}
dark\_channel(R,G,B) = \min \{ R,G,B\}
\end{equation}

For simplicity, we define
\begin{equation}
\begin{array}{*{20}{c}}
{\begin{array}{*{20}{c}}
{{X_1} = DWT(LQ),{X_2} = dark\_channel(LQ)}\\
{{Y_1} = DWT(HQ),{Y_2} = dark\_channel(HQ)}
\end{array}}\\
{X = \{ {X_1},{X_2}\} ,Y = \{ {Y_1},{Y_2}\} }
\end{array}
\end{equation}

In consideration of the two above aspects, we propose a novel training method. On one hand, four wavelet subbands and dark channel of low-quality image pass through the same convolutional layers:
\begin{equation}
\hat Y = f(\{ {X_1},{X_2}\} ) + \{ {X_1},{X_2}\}
\end{equation}

Where $\{\cdot , \cdot\}$ means concatenating the two tensors to a tensor and $f( \cdot )$ represents the convolutional architecture. In this paper we use dense-net proposed in~\cite{huang2016densely}, which is better to represent the features of the image and obtains the faster convergence.

On the other hand, the loss function of these two results are evaluated separately. As for the first task, we should make the wavelet transform of rain image $Y_{1}$ restore to this of ground truth ${\hat Y}_1$under the criteria of Frobenius norm. We set the loss $L_1$ to:
\begin{equation}
{L_1} = \frac{1}{N}\sum\limits_{i = 1}^N {\left\| {{Y_{1i}} - {{\hat Y}_{1i}}} \right\|_F^2}
\end{equation}

Meanwhile, we make the dark channel feature of the rain image $Y_2$ and this feature of ground truth ${\hat Y}_2$ as close as possible, which is useful to detect the area of haze veil and remove it. The loss $L_2$ is set:

\begin{equation}
{L_2} = \frac{1}{N}\sum\limits_{i = 1}^N {\left\| {{Y_{2i}} - {{\hat Y}_{2i}}} \right\|_F^2}
\end{equation}

We combine above two loss functions, $L_1$ and $L_2$, and obtain the final goal of optimization $L_{total}$ which is:

\begin{equation}
  L_{total} = L_1 + \alpha L_2
\end{equation}

Where $\alpha$  plays the balance role of $L_1$ and $L_2$, we empirically set $\alpha$ to 0.5 in this paper because we find that the result of our method is insensitive to the different value of $\alpha$ in a large range. The parameters in this model are optimized by back-propagation. After DJRHR-net has been trained, the whole rain removal process is similar to last section:

Firstly, we generate the wavelet subbands and dark channel of a rain image.
\begin{equation}
{X_1} = DWT(LQ),{X_2} = dark\_channel(LQ)
\end{equation}

Then, we concatenate the $X_1$ and $X_2$ to a tensor with 13 channels and pass it through the trained residual network(DJRHR-net).
\begin{equation}
\{ {\hat Y_1},{\hat Y_2}\}  = f(\{ {X_1},{X_2}\} ) + \{ {X_1},{X_2}\}
\end{equation}

At last, the inverse wavelet transform is used to generate final high-quality result.
\begin{equation}\label{idwt}
HQ = IDWT({\hat Y_1})
\end{equation}

We have to emphasize that the dark channel feature in this model is just used for removing the haze in an indirect way. In training process, the dark channel error is used for updating the weights of convolutional network. But in test process, considering that the train result of dark channel feature is different from the dark channel of wavelet subbands, we preserve the wavelet subbands and discard dark channel, as Equation~\ref{idwt} shows.

\section{Experiment}\label{experiments}

In last section, we propose two networks, SRR-net and DJRHR-net, which process the sparse and dense rain respectively. To evaluate the performance of our method, we use both the synthetic test data and the real-world images to compare our approch with two recent state-of-the-art deraining methods based on network, which contains Detail-net~\cite{Fu2017Removing}, JORDER~\cite{yang2016joint} and SIRR-net\cite{Fu2016Clearing}, which removes the final enhancement for fair.

\subsection{Dataset generation}

For learning the parameters of the SRR-net and DJRHR-net, we construct two datasets to deal with different situations. As for the rain images without the haze veil, we simply make use of the dataset from~\cite{Fu2017Removing} as ground truth and add 12 types of rain streaks~\cite{li2016rain} to obtain \textbf{TrainSet A}.

\begin{figure}[ht]
  \centering
  \includegraphics[width=0.45\textwidth]{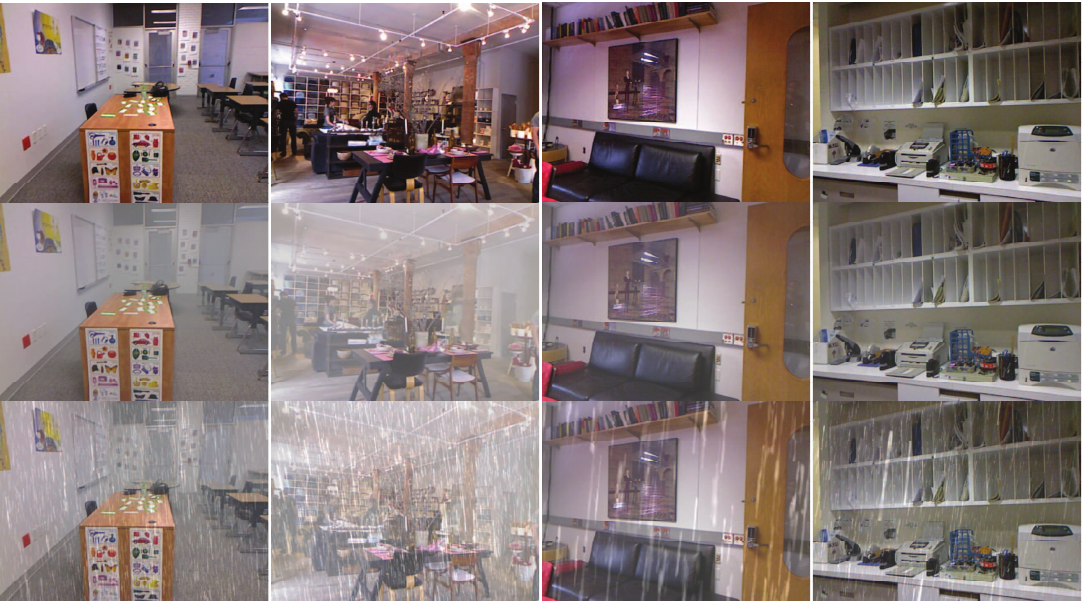}\\
  \caption{Some images in TrainSet B. The top row is the ground truth in NYU dataset, the middle row is the image with haze veil and the bottom row is the final image with rain and haze.}\label{TrainSetB}
\end{figure}

Furthermore, in order to train the parameters of our DJRHR-net, we create a new dataset as \textbf{TrainSet B}, which contains a number of low-quality(LQ) and high-quality(HQ) image pairs with rain and haze veil noises. In view of the fact that the formation of haze is based on the depth information of the images, we firstly select 1449 RGBD images from the NYU Depth Dataset V2~\cite{Silberman:ECCV12} and generate the haze according to the atmospheric scattering model. Next, we also increase the 12 types of rain streaks~\cite{li2016rain} to these foggy images. Figure~\ref{TrainSetB} shows the part of the TrainSet B.

Besides the experiment on our own synthetic dataset, TrainSet A and TrainSet B, we also choose real-world rain data~\cite{li2016rain} to evaluate our method.
\subsection{Training setup}

For the SRR-net, we simply set the depth of the network to 20. We spend about 8 hours on training the SRR-net by using the Caffe~\cite{jia2014caffe} and use Adam with weight decay of $10^{-6}$ and mini-batch size of 64. For DJRHR-net, we remove the batch normalization and pooling layer to get better regression effect. Besides, we set the growth rate $K$ to 12, the number of the denseblocks $L$ is 3. We use the pytorch to construct the network and use Adam with weight decay of $10^{-4}$ and mini-batch size of 10. We start with a learning rate of $10^{-3}$ and the learning rate decay of 0.95.

\subsection{Experiment on synthetic rain data}

Figure~\ref{fig:synthesized} shows the visual comparison for several methods on synthesized rain images. As we can see, the results of SIRR-net~\cite{Fu2016Clearing}, Detail-net~\cite{Fu2017Removing} and JORDER~\cite{yang2016joint} look unnatural and remove the rain streaks badly, while our method achieves better performance.

Considering that the ground truth is known for the synthetic test data, we use PSNR, SSIM~\cite{wang2004image} and NIQE~\cite{mittal2013making} for a quantitative evaluation. A higher PSNR or SSIM indicates that the image is closer to the ground truth, but the lower NIQE means a higher image quality. All the best results are boldfaced. As shown in Table~\ref{table:SyntheticQuantitativeResult}, our SRR-net obtains higher PSNR/SSIM and lower NIQE average than other methods for 200 test images.

\subsection{Experiment on real-world rain data}

Figure~\ref{fig:realworld} also shows the results of several state-of-the-art methods on the real-word images. As shown in each row, our method DJRHR-net always achieves better performance than others in the aspect of rain and haze veil removal. As for the heavy rain, DJRHR-net is valid to remove these noises.

\subsection{Study of SRR-net and DJRHR-net parameters}

The number of the denseblocks $L$ and the growth rate $K$ are the main hyper-parameters in our DJRHR-net. As shown in Table~\ref{Tab:average PSNR}, we know that the deeper structure can improve the learning ability. For the better performance, we set the $K=12$ and $L=3$ for our experiments above.
\begin{table}[htbp]
\centering
\caption{Average PSNR for different network parameters}
\label{Tab:average PSNR}
\begin{tabular}{l|ccc}
\hline
 &$K=8$ &$K=10$ &$K=12$\\ \hline  % \hline 在此行下面画一横线
$L =1$  &14.85 &14.90 &15.40\\         % \\ 表示重新开始一行
$L =2$  &15.10 &15.20 &15.69\\        % & 表示列的分隔线
$L =3$  &15.50 &15.55 &15.75\\ \hline
\end{tabular}
\end{table}

\section{Conclusion}

In this paper, we propose a novel convolutional neural network based on wavelet and dark channel. Considering that rain streaks correspond to high frequency component of the image, we attempt to use wavelet transform to separate the rain streaks and background. More specifically, HL, LH of the rain image are more inclined to represent the raindrops and the edges of the ground truth respectively. However, the dense rain makes the image look like haze veil. So we extract dark channel as a feature map in network, which plays an important role in removing the haze veil. Finally, we design two architectures, SRR-net and DJRHR-net to process the sparse and dense rain streaks respectively and test our model on both synthetic and real-world images, all of which obtain very impressive performance.

\bibliographystyle{IEEEtran}
\bibliography{ICPRbib}
\end{document}